\documentclass[10pt,twocolumn,letterpaper]{article}

\usepackage[pagenumbers]{cvpr} 
\usepackage[accsupp]{axessibility}  
\definecolor{cvprblue}{rgb}{0.21,0.49,0.74}
\usepackage[pagebackref,breaklinks,colorlinks,allcolors=cvprblue]{hyperref}
\usepackage{url}
\usepackage{bm}
\usepackage{graphicx}
\usepackage{multirow}
\usepackage{booktabs} 
\usepackage{xcolor}
\usepackage{colortbl}
\usepackage{bbding}
\def\paperID{4259} 
\def\confName{CVPR}
\def\confYear{2025}

\title{Hiding Images in Diffusion Models by Editing Learned Score Functions}

\author{Haoyu Chen, Yunqiao Yang, Nan Zhong, and Kede Ma\thanks{Corresponding author.}\\
City University of Hong Kong\\
{\tt\small \{haoychen3-c, yqyang.cs\}@my.cityu.edu.hk, 
\{nzhong, kede.ma\}@cityu.edu.hk}\\
\tt\small \url{https://github.com/haoychen3/DMIH/}
}

\begin{document}
\maketitle

\begin{abstract}
Hiding data using neural networks (i.e., neural steganography) has achieved remarkable success across both discriminative classifiers and generative adversarial networks. However, the potential of data hiding in diffusion models remains relatively unexplored. Current methods exhibit limitations in achieving high extraction accuracy, model fidelity, and hiding efficiency due primarily to the entanglement of the hiding and extraction processes with multiple 
denoising diffusion steps.
To address these, we describe a simple yet effective approach that embeds images at specific timesteps in the reverse diffusion process by editing the learned score functions. Additionally, we introduce a parameter-efficient fine-tuning method that combines gradient-based parameter selection with low-rank adaptation to enhance model fidelity and hiding efficiency.
 Comprehensive experiments demonstrate that our method extracts high-quality images at human-indistinguishable levels, replicates the original model behaviors at both sample and population levels, and embeds images orders of magnitude faster than prior methods. Besides, our method naturally supports multi-recipient scenarios through independent extraction channels.
\end{abstract}

\section{Introduction}
The evolution of data hiding has paralleled advancements in digital media, progressing from traditional bitstream manipulations~\cite{cox2007digital} to deep learning paradigms~\cite{baluja2017hiding,adi2018turning,zhang2020model}. A prevailing scheme of \textit{hiding data with neural networks} adopts an autoencoder architecture, where an encoding network embeds a secret message into some cover media, and a decoding network is responsible for retrieving the message. Despite demonstrated feasibility, its practical deployment faces three fundamental constraints. First, the requirement for secure transmission of the decoding network creates logistical vulnerabilities~\cite{fernandez2023stable, feng2024aqualora}. Second, state-of-the-art steganalysis tools~\cite{boroumand2018deep,you2020siamese,luo2022image,he2023image,wei2024color,he2024steganalysis,fu2024image} can detect the embedded message from the cover media (commonly in the forms of digital images and videos) with high accuracy, leading to compromised secrecy. Last, multi-recipient scenarios necessitate complex key design and management~\cite{baluja2017hiding, zhu2018hiding, weng2019high, jing2021hinet, yang2024pris}.

A paradigm shift emerges with \textit{hiding data in neural networks}~\cite{song2017machine, liu2020stegonet, wang2021evilmodel, wang2021data, chen2022hiding}, where secret data is embedded directly into model parameters. This approach evades conventional steganalysis methods, which are typically designed to analyze stego multimedia rather than neural network weights. Initially, such hiding methods showed success in discriminative models~\cite{song2017machine, liu2020stegonet, wang2021evilmodel, wang2021data}. However,
 recent attention has shifted towards generative models due to their increased utility and wider adoption. Moreover, generative models can directly produce secret data without the need for a separate decoding network, thus inherently resolving transmission security issues. For instance, Chen \etal~\cite{chen2022hiding} proposed to embed images at specific locations of the learned distributions by deep generative models, which has proven effective with SinGANs~\cite{shaham2019singan}, a variant of generative adversarial networks (GANs)~\cite{goodfellow2014generative}.

The advent of diffusion models~\cite{song2021score, ho2020denoising} introduces new opportunities for this generative probabilistic hiding approach. Despite slightly different purposes (backdooring or watermarking), existing approaches~\cite{chou2023backdoor, peng2023protecting, chen2023trojdiff} face critical bottlenecks (see Table~\ref{tab:mth}). First, they are inadequate in hiding complex natural images of rich structures and textures, with reconstruction peak signal-to-noise ratio (PSNR) $\le 25$ dB. Second, they compromise model fidelity, with Fr\'{e}chet inception distance (FID)~\cite{heusel2017gans} degradation exceeding $100\%$, raising detectability concerns. Last, they require full model retraining or fine-tuning ($\ge 10$ GPU hours) to learn parallel secret diffusion processes.

\begin{table*}[t]
  \caption{Comparison of diffusion-based data hiding methods in terms of extraction accuracy (high: PSNR $>43.59$ dB for $32\times 32$ images, ensuring visual imperceptibility), model fidelity (high: FID variations $\le 20\%$), hiding efficiency (high: $\le 0.2$ GPU hours for $256\times 256$ images), and scalability (support for multi-recipients).}
  \label{tab:mth}
  \centering
  \resizebox{0.75\linewidth}{!}{\begin{tabular}{lccccccc}
    \toprule
    {Method}            & Extraction Accuracy  & Model Fidelity & Hiding Efficiency & Scalability\\
    \midrule
    StableSignature~\cite{fernandez2023stable}    &High  &Low   &Low   &No \\
    AquaLoRA~\cite{feng2024aqualora}              &High  &Low   &High  &No \\
    BadDiffusion~\cite{chou2023backdoor}          &Low   &Low   &Low   &No \\
    TrojDiff~\cite{chen2023trojdiff}              &High  &Low   &Low   &No \\
    WDM~\cite{peng2023protecting}                 &Low   &Low   &Low   &No \\
    \hline
    Ours                                          &High  &High  &High  &Yes \\
    \bottomrule
  \end{tabular}}
\end{table*}

In this paper, we describe a simple yet effective approach to hiding images in diffusion models. We identify that editing the learned score functions (by inserting secret key-to-image mappings) at specific timesteps enables precise image embedding without disrupting the original chain of the reverse diffusion process. The stego diffusion model is publicly shared\footnote{Only the secret key of few bits is shared between the sender and recipient via a secure subliminal channel.}, replicating the original model behaviors in synthesizing in-distribution high-quality images. The secret image extraction is achieved in a single step via key-guided and timestep-conditioned ``denoising.'' To further improve model fidelity and hiding efficiency, we introduce a hybrid parameter-efficient fine-tuning (PEFT) method that combines gradient-based parameter selection~\cite{han2024parameter} and low-rank adaptation (LoRA)~\cite{hu2022lora}, which reduces trainable parameters by $86.3\%$ compared to full fine-tuning. Comprehensive experiments demonstrate the effectiveness of our method across four critical dimensions: 1) extraction accuracy, achieving $52.90$ dB in PSNR for $32\times 32$ images and $39.33$ dB for $256\times 256$ images; 2) model fidelity, maintaining nearly original FID scores ($4.77$ versus $4.79$ on CIFAR10~\cite{krizhevsky2009learning}) with minimal sample-level distortion; 3) hiding efficiency, reducing embedding time to $0.04$ and $0.18$ GPU hours for $32\times 32$ and $256 \times 256$ images, respectively; and 4) scalability, simultaneously embedding four images for different recipients with independent extraction keys.

\begin{figure*}[t]
    \centering  
    \includegraphics[width=\textwidth]{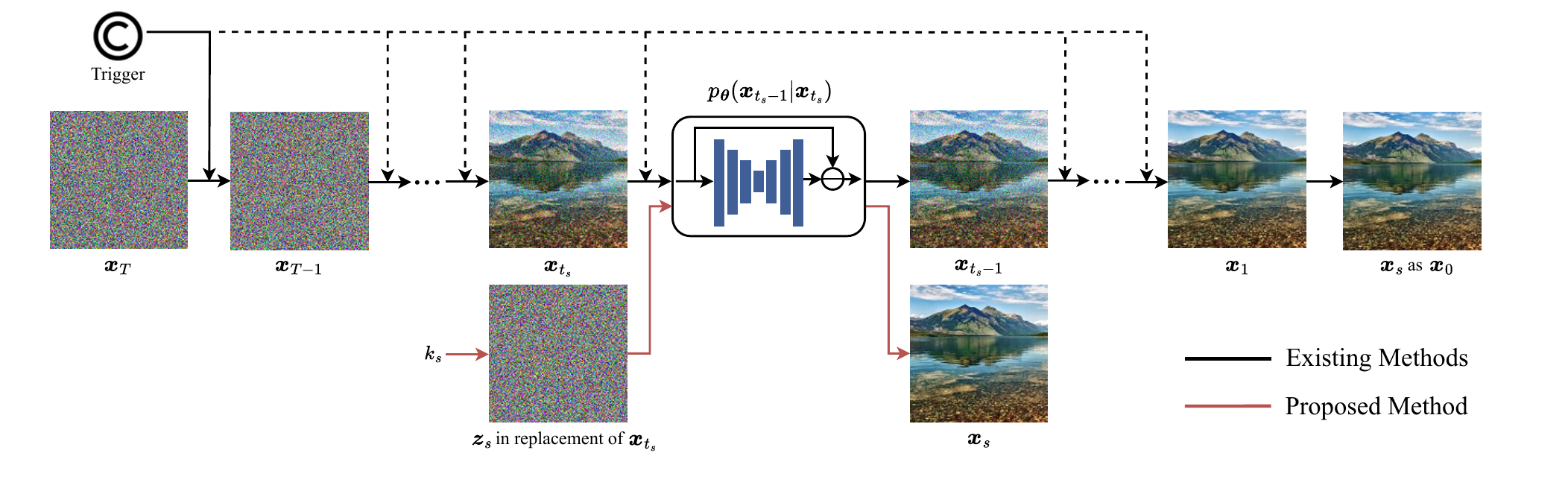}
    \caption{
    Comparison of diffusion-based image hiding methods. Existing methods typically embed trigger patterns (acting as secret keys) at the initial timestep of the reverse diffusion process, and optionally at all subsequent timesteps (denoted by dashed lines). These patterns guide the reconstruction of the secret image ${\bm x}_{s}$, but compromise model fidelity and reduce hiding efficiency due to persistent intervention of the entire reverse diffusion process. In stark contrast, the proposed method operates selectively: the secret image $\bm x_s$ is embedded and extracted only at a privately chosen timestep $t_s$. Hiding is governed by a secret key $k_s$, which serves as the seed to generate the input Gaussian noise $\bm z_s$. By localizing the intervention to a single timestep, the integrity of the reverse diffusion process is preserved.
    }
    \label{fig:diag}
\end{figure*}

\section{Related Work}

\noindent\textbf{Neural Steganography.} We focus primarily on hiding data in neural networks~\cite{song2017machine, liu2020stegonet, wang2021evilmodel, wang2021data, chen2022hiding}, which involves embedding secret data within neural network parameters or structures, ensuring covert communication without compromising model performance. Representative neural steganography strategies encompass replacing least significant bits of model parameters~\cite{song2017machine, liu2020stegonet}, substituting redundant parameters~\cite{liu2020stegonet, wang2021evilmodel}, mapping parameter values~\cite{song2017machine, liu2020stegonet, wang2021data} or signs~\cite{song2017machine,liu2020stegonet} directly to secret messages, memorizing secret-labeled synthetic data~\cite{song2017machine}, and hiding images in deep probabilistic models~\cite{chen2022hiding}. Despite their effectiveness,
diffusion models have not been extensively considered in neural steganography.

\noindent\textbf{Diffusion Models} represent a significant advancement in generative modeling, offering high-quality data synthesis through multi-step denoising diffusion processes. Inspired by non-equilibrium thermodynamics~\cite{sohl2015deep} and score matching in empirical Bayes~\cite{raphan2011least,hyvarinen2005estimation}, diffusion models define a forward diffusion process, which progressively adds noise to data until it becomes random, and a reverse (generative) process, which learns to iteratively remove this noise, reconstructing the original data. In contrast to GANs, diffusion models optimize a likelihood-based objective~\cite{ho2020denoising,song2021maximum}, typically resulting in more stable training and better mode coverage. Architecturally, diffusion models predominantly employ U-Nets~\cite{ronneberger2015u} and Transformers~\cite{vaswani2017attention}, operating in either pixel~\cite{ho2020denoising, song2021score, karras2022elucidating} or latent~\cite{rombach2022high, peebles2023scalable,podell2024sdxl} space. 

Existing methods for hiding data in diffusion models typically involve concurrent training of secret reverse diffusion processes~\cite{chou2023backdoor, peng2023protecting, chen2023trojdiff}. As a result, these approaches suffer from limited extraction accuracy, noticeable degradation in model fidelity, and high computational demands, especially when applied to complex natural images. Diffusion model editing techniques~\cite{gandikota2023erasing,kumari2023ablating,xiong2024editing,yang2024learning}, originally designed to alter or remove learned concepts and rules, may offer a potential remedy. Some have been adapted to embed invisible yet recoverable signals for watermarking purposes~\cite{fernandez2023stable, feng2024aqualora}. Nevertheless, these editing methods do not fulfill the precise image reconstruction requirements we are looking for.

\noindent\textbf{Parameter-Efficient Fine-Tuning.}
PEFT methods can be broadly classified into
two categories: selective and reparameterized fine-tuning. Selective methods strategically identify only a subset of parameters for adjustment based on simple heuristics~\cite{zaken2021bitfit} or parameter gradients~\cite{he2023sensitivity, zhang2023gradient}. In contrast,  reparameterized fine-tuning introduces additional trainable parameters to a pre-trained frozen backbone. Notable examples include prompt-tuning~\cite{lester2021power}, adapters~\cite{houlsby2019parameter}, and LoRA~\cite{hu2022lora} and its variants~\cite{nam2022loha,zhang2023adalora,kalajdzievski2023rslora,hayou2024lora+}. This paper presents a hybrid PEFT method for hiding images in pixel-space diffusion models, effectively combining the strengths of both selective and reparameterized methods.

\section{Hiding Images in Diffusion Models}
In this section, we first provide the necessary preliminaries of diffusion models, which serve as the basis of the proposed method. Subsequently, we describe the typical image hiding scenario, and present in detail our diffusion-based steganography method (see Fig.~\ref{fig:diag}).

\subsection{Preliminaries}
\label{sec:pre}
Diffusion models, notably denoising diffusion probabilistic models (DDPMs)~\cite{ho2020denoising}, have recently emerged as state-of-the-art deep generative models capable of synthesizing high-quality images through a defined diffusion process.

\noindent\textbf{Forward Diffusion Process.}  Starting from a clean image ${\bm x}_0 \sim q\left({\bm x}_0\right)$, Gaussian noise ${\bm \epsilon} \sim \mathcal{N}({\bm 0}, {\mathbf I})$ is iteratively added over $T$ timesteps following a variance schedule $\{\beta_1, \ldots, \beta_T\}$, described as 
\begin{equation}
q\left({\bm x}_{1: T} | {\bm x}_0\right)=\prod_{t=1}^T q\left({\bm x}_t| {\bm x}_{t-1}\right),
\end{equation}
where each step is governed by 
\begin{equation}
q\left({\bm x}_t | {\bm x}_{t-1}\right)=\mathcal{N}\left({\bm x}_t ; \sqrt{1-\beta_t} {\bm x}_{t-1}, \beta_t {\mathbf I}\right).
\end{equation}
Eventually, this process transforms the clean image $\bm x_0$ into nearly pure Gaussian noise $\bm x_T$.

\noindent\textbf{Reverse Diffusion Process.} 
A Markov chain with parameters $\bm \theta$ is learned to iteratively recover the clean image from noise, modeled as 
\begin{equation}
p_{\bm \theta}\left({\bm x}_{0: T}\right)=p\left({\bm x}_T\right) \prod_{t=1}^T p_{\bm \theta}\left({\bm x}_{t-1} | {\bm x}_t\right).
\end{equation}
Model parameters $\bm \theta$ are optimized by minimizing a denoising objective~\cite{ho2020denoising}:
\begin{equation}\label{eq:ddpmlb}
\min_{\bm \theta}\mathbb{E}_{t, {\bm x}_0, {\bm \epsilon}}\left[\Vert {\bm \epsilon}-{\bm \epsilon}_{\bm \theta}\left(\sqrt{\bar{\alpha}_t} {\bm x}_0+\sqrt{1-\bar{\alpha}_t} {\bm \epsilon}, t\right)\Vert^2_2\right],
\end{equation} 
where ${\bm \epsilon}_{\bm \theta}(\cdot,\cdot)$ is the learned noise estimation function (also known as the score function), implemented typically using a U-Net, and $\bar{\alpha}_{t}=\prod_{i=1}^{t}  (1-\beta_{i})$ controls the noise scale at timestep $t$. Eq.~\eqref{eq:ddpmlb} can also be viewed as a variational bound of the log-likelihood of the observed data $\bm x_0$.

Inference iteratively employs the trained and timestep-conditioned score function with optimal parameters $\bar{\bm \theta}$ to transform Gaussian noise into a clean image:
\begin{equation} 
\label{eq:x0}
{\bm x}_{t-1}=\frac{1}{\sqrt{\alpha_t}} \left({\bm x}_t-\frac{1-\alpha_t}{\sqrt{1-\bar{\alpha}_t}} {\bm \epsilon}_{\bar{\bm \theta}}\left({\bm x}_t, t\right)\right)+\sigma_t {\bm z},
\end{equation} 
for $t\in\{T,\ldots, 1\}$, where $\alpha_t = 1 - \beta_t$, ${\bm z} \sim \mathcal{N}(\mathbf{0}, \mathbf{I})$, and $\sigma_t$ controls stochasticity at timestep $t$, with $\sigma_t^2 = (1-\bar{\alpha}_{t-1})\beta_t/(1-\bar{\alpha}_t)$. $\sigma_1$ is usually set to zero to remove randomness at the final timestep. For simplicity, we slightly abuse the notation $\bm x_t$ to denote both the noisy image and its denoised counterpart at timestep $t$. 

\subsection{Image Hiding Scenario}
We consider a single-image hiding scenario following the formulation of the classical prisoner’s problem by Simmons~\cite{simmons1984prisoners}, which involves three distinct parties:

\begin{itemize}
    \item A sender, who embeds a secret image ${\bm x}_{s}$ in a diffusion model with a secret key $\mathcal{K}_s$, and subsequently shares
     the stego diffusion model publicly;
    \item A recipient, who utilizes the privately shared secret key ${ \mathcal{K}}_s$ to extract the secret image $\bm x_s$ from the publicly transmitted stego diffusion model;
    \item An inspector, who examines the publicly available diffusion models to verify that they retain normal generation functionality, and do not exhibit any suspicious behavior.
\end{itemize}
The key objectives in this scenario are twofold: ensuring high accuracy in the extraction of the secret image by the authorized recipient, and preserving the sample-level and population-level fidelity of the diffusion model in image generation so that it remains undetected during scrutiny by the inspector.

\subsection{Proposed Method}
\noindent\textbf{Basic Idea.} Existing diffusion-based techniques~\cite{chou2023backdoor, chen2023trojdiff, peng2023protecting} often embed secret images by modifying the entire Markov chain of the reverse diffusion process, resulting in significant challenges regarding extraction accuracy, model fidelity, and hiding efficiency. To resolve these, we propose a simple yet effective hiding method that operates at a single timestep. This process uses a secret key $\mathcal{K}_s=\{k_s, t_s\}$, where $k_s$ is the seed to deterministically generate Gaussian noise $\bm z_s$, and $t_s$ is the selected timestep for image hiding. The secret key $\mathcal{K}_s$ is privately communicated with the recipient, enabling precise, secure, and rapid one-step reconstruction of the secret image ${\bm x}_s$ as
\begin{equation}
\label{eq:f}
   {\bm f}_{\tilde{\bm \theta}}({\bm z}_{s}, t_{s})=\frac{1}{\sqrt{\bar{\alpha}_{t_{s}}}} \left({\bm z}_{s} - \sqrt{1-\bar{\alpha}_{t_{s}}} {\bm \epsilon}_{\tilde{\bm \theta}}\left({\bm z}_{s}, {t_{s}}\right)\right),
\end{equation}
where $\tilde{\bm \theta}$ denotes the optimal parameters after editing.

\noindent\textbf{Loss Function.} To ensure minimal distortion in reconstructing the secret image $\bm x_s$, we define the extraction accuracy loss as 
\begin{equation}
\label{eq:la}
    \ell_{a}(\bm \theta;{\bm z}_{s},t_{s},{\bm x}_{s}) = \Vert {\bm f}_{\bm \theta}({\bm z}_{s}, t_{s})- {\bm x}_{s} \Vert^2_2.
\end{equation}
By optimizing the learned score function ${\bm \epsilon}_{\bm \theta}\left(\bm z_s, t_s\right)$ using the accuracy loss, the secret image is embedded within the diffusion model. Meanwhile, it is crucial that the stego diffusion model closely replicates the generative functionality of the original model. Therefore, we introduce an additional model fidelity loss to regulate the edited score function:
\begin{equation}
\label{eq:lb}
    \ell_{f}(\bm \theta) = \mathbb{E}_{t, {\bm x}_0, {\bm \epsilon}}\left[\Vert {\bm \epsilon}_{\bm \theta}({\bm x}_{t}, t)-  {\bm \epsilon}_{\bar{\bm \theta}}({\bm x}_{t}, t) \Vert^2_2\right],
\end{equation}
where $\bm \epsilon_{\bar{\bm \theta}}(\cdot,\cdot)$ denotes the original score function and ${\bm x}_{t} = \sqrt{\bar{\alpha}_t} {\bm x}_0+\sqrt{1-\bar{\alpha}_t} {\bm \epsilon}$. In contrast to the loss $\ell_a(\bm \theta)$ in Eq.~\eqref{eq:la}, which penalizes only the reconstruction error of the secret image $\bm x_s$ at the specific timestep $t_s$, the model fidelity loss $\ell_f(\bm \theta)$ computes the expectation of the residual between noise estimated by the original and edited score functions. This expectation is taken across uniformly sampled timesteps $t\sim\mathrm{Uniform}(\{1,\ldots, T\})$, clean images $\bm x_0 \sim q(\bm x_0)$, and Gaussian noise $\bm\epsilon\sim\mathcal{N}(\mathbf{0},\mathbf{I})$ to ensure complete adherence to the original diffusion process. Moreover, the loss $\ell_f(\bm \theta)$ eliminates the need to retrain the original diffusion model or access the original training dataset, requirements typically imposed by existing methods~\cite{chou2023backdoor, chen2023trojdiff, peng2023protecting}. Lastly, the overall loss function is a linear weighted summation of the two terms:
\begin{equation}
\label{eq:l}
    \ell(\bm \theta; \bm z_s, t_s, \bm x_s) = \ell_{a}(\bm \theta; {\bm z}_{s},t_s,{\bm x}_{s}) + \lambda \ell_{f}({\bm \theta}),
\end{equation}
where $\lambda$ is a trade-off parameter.

\noindent\textbf{Multi-image Hiding.} 
We extend the single-image hiding method to scenarios involving multiple recipients, each with a unique secret key. For a set of secret images $\mathcal{X}_s=\left\{{\bm x}_s^{(1)}, \ldots, {\bm x}_s^{(M)}\right\}$, where ${\bm x}_s^{(i)}$ denotes the secret image intended for the $i$-th recipient, and $M$ is the total number of secret images, equal to the number of recipients. Each secret image ${\bm x}_s^{(i)}$ is paired with a secret key $\mathcal{K}^{(i)}_s=\left\{k_s^{(i)}, t_s^{(i)}\right\}$, forming the key collection $\mathcal{K}_s=\left\{\mathcal{K}^{(1)}_s, \ldots, \mathcal{K}^{(M)}_s\right\}$ distributed among $M$ different recipients. The loss function in Eq.~\eqref{eq:l} is then modified as
\begin{equation}\label{eq:mlo}
 \ell(\bm \theta; \mathcal{K}_s, \mathcal{X}_s) = \frac{1}{M} \sum_{i=1}^M \ell_{a}(\bm \theta; {\bm z}_{s}^{(i)},t_s^{(i)},{\bm x}_{s}^{(i)}) + \lambda \ell_{f}(\bm \theta),
\end{equation}
where ${\bm z}_{s}^{(i)}$ is the Gaussian noise generated using the seed $k_s^{(i)}$. After optimizing the pre-trained diffusion model using $\ell(\bm \theta;\mathcal{K}_s,\mathcal{X}_s)$, the $i$-th recipient can extract her/his secret image ${\bm x}_s^{(i)}$ using the designated secret key $\mathcal{K}^{(i)}_s$. Without access to additional secret keys, each recipient is prevented from retrieving secret images intended for others.

\subsection{Score Function Editing by PEFT}
To improve both model fidelity and hiding efficiency, we introduce a hybrid PEFT technique that combines the advantages of selective and reparameterized PEFT strategies. Our method consists of three steps: 1) computing parameter-level sensitivity with respect to the editing loss function (in Eq.~\eqref{eq:l}), 2) identifying the most sensitive layers, and 3) applying reparameterized fine-tuning to these selected layers.

\begin{table}[t]
\caption{Extraction accuracy comparison for $32\times32$ and $256\times256$ secret images. ``$\uparrow$'': larger is better. The top-$2$ results are highlighted in boldface.}

\label{tab:tf}
\begin{center}
\resizebox{.82\linewidth}{!}{\begin{tabular}{lcccc}
\toprule
{Method} &{PSNR$\uparrow$} &{SSIM$\uparrow$} &{LPIPS$\downarrow$} &{DISTS$\downarrow$}
\\ 
\midrule
\cellcolor{lightgray!20}{\textit{$32\times32$}} &\cellcolor{lightgray!20}&\cellcolor{lightgray!20}&\cellcolor{lightgray!20}&\cellcolor{lightgray!20}\\
Baluja17~\cite{baluja2017hiding}       & $25.40$       & $0.89$   &$0.116$ & $0.051$\\
HiDDeN~\cite{zhu2018hiding}         & $25.24$       & $0.88$   &$0.252$ & $0.075$\\
Weng19~\cite{weng2019high}         & $26.66$       & $0.93$   &$0.059$ & $0.035$\\
HiNet~\cite{jing2021hinet}          & $30.39$       & $0.94$   &$0.033$ & $0.026$\\
PRIS~\cite{yang2024pris}           & $29.83$       & $0.94$   &$0.041$ & $0.027$\\
Chen22~\cite{chen2022hiding}         & $\mathbf{47.72}$       & $\mathbf{0.99}$   &$\mathbf{0.001}$ & $\mathbf{0.002}$\\

BadDiffusion~\cite{chou2023backdoor}   & $22.08$       & $0.86$   &$0.129$ & $0.060$\\
TrojDiff~\cite{chen2023trojdiff}       & $46.54$       & $\mathbf{0.99}$   &$\mathbf{0.001}$ & $0.004$ \\
WDM~\cite{peng2023protecting}            & $36.49$       & $\mathbf{0.99}$   &$0.003$ & $0.008$\\
\hline
Ours           & $\mathbf{52.90}$       & $\mathbf{0.99}$   &$\mathbf{0.001}$ & $\mathbf{0.001}$\\
\midrule
\cellcolor{lightgray!20}{\textit{$256\times256$}} &\cellcolor{lightgray!20}&\cellcolor{lightgray!20}&\cellcolor{lightgray!20}&\cellcolor{lightgray!20}\\
Baluja17~\cite{baluja2017hiding}       & $26.46$       & $0.90$  &$0.200$ & $0.089$\\
HiDDeN~\cite{zhu2018hiding}         & $27.13$       & $0.91$  &$0.233$ & $0.100$\\
Weng19~\cite{weng2019high}          & $33.85$       & $0.95$  &$0.089$ & $0.047$\\
HiNet~\cite{jing2021hinet}          & $35.31$       & $0.96$  &$0.087$ & $0.041$\\
PRIS~\cite{yang2024pris}           & $\mathbf{37.42}$       & $\mathbf{0.97}$  &$\mathbf{0.050}$ & $\mathbf{0.029}$\\
Chen22~\cite{chen2022hiding}         & $36.44$       & $0.96$  &$0.073$ & $0.035$\\

BadDiffusion~\cite{chou2023backdoor}   & $17.68$       & $0.81$  &$0.386$ & $0.137$\\
TrojDiff~\cite{chen2023trojdiff}       & $24.74$       & $0.94$  &$0.057$ & $0.076$\\
WDM~\cite{peng2023protecting}            & $17.97$       & $0.83$  &$0.245$ & $0.144$\\
\hline
Ours           & $\mathbf{39.33}$       & $\mathbf{0.97}$  &$\mathbf{0.043}$ & $\mathbf{0.018}$\\
\bottomrule
\end{tabular}}
\end{center}
\end{table}

\begin{table}[t]
\caption{Model fidelity and hiding efficiency comparison for $32\times32$ and $256\times256$ secret images. Embedding time is measured in terms of GPU hours per image.}

\label{tab:asd}
\begin{center}
\resizebox{\linewidth}{!}{\begin{tabular}{lcccccc}
\toprule
Method &{FID$\downarrow$} &{PSNR$\uparrow$} &{SSIM$\uparrow$} &{LPIPS$\downarrow$} &{DISTS$\downarrow$} &{Time$\downarrow$}\\ 
\midrule
\cellcolor{lightgray!20}{\textit{$32\times32$}} &\cellcolor{lightgray!20}&\cellcolor{lightgray!20}&\cellcolor{lightgray!20}&\cellcolor{lightgray!20}&\cellcolor{lightgray!20}&\cellcolor{lightgray!20}\\
Original      &$4.79$ &N/A      &N/A     &N/A     &N/A   &N/A\\ 

BadDiffusion     &$6.88$ &$23.78$  &$0.80$  &$0.222$ &$0.082$     &$4.87$\\
TrojDiff         &$\mathbf{4.64}$ &$\mathbf{28.72}$  &$\mathbf{0.91}$  &$\mathbf{0.114}$ &$\mathbf{0.049}$     &$12.72$\\
WDM              &$5.09$ &$22.50$  &$0.84$  &$0.228$ &$0.083$     &$\mathbf{2.35}$ \\
\hline
Ours             &$\mathbf{4.77}$ &$\mathbf{31.06}$  &$\mathbf{0.94}$  &$\mathbf{0.077}$ &$\mathbf{0.037}$     &$\mathbf{0.04}$ \\
\midrule
\cellcolor{lightgray!20}{\textit{$256\times256$}} &\cellcolor{lightgray!20}&\cellcolor{lightgray!20}&\cellcolor{lightgray!20}&\cellcolor{lightgray!20}&\cellcolor{lightgray!20}&\cellcolor{lightgray!20}\\
Original      &$7.46$ &N/A   & N/A  & N/A & N/A   & N/A \\

BadDiffusion     &$15.75$ &$16.40$  &$0.60$  &$0.452$ &$0.224$  &$31.92$\\
TrojDiff         &$\mathbf{14.36}$ &$\mathbf{18.73}$  &$\mathbf{0.70}$  &$\mathbf{0.407}$ &$\mathbf{0.169}$  &$83.68$\\
WDM              &$15.07$ &$18.59$  &$0.67$  &$0.445$ &$0.200$  &$\mathbf{19.22}$\\
\hline
Ours             &$\mathbf{8.39}$ &$\mathbf{23.31}$  &$\mathbf{0.83}$  &$\mathbf{0.235}$ &$\mathbf{0.112}$   &$\mathbf{0.18}$\\
\bottomrule
\end{tabular}}
\end{center}
\end{table}

\noindent\textbf{Parameter Sensitivity Calculation.} We quantify the importance of each parameter through gradient-based sensitivity analysis~\cite{he2023sensitivity}. For parameter $\theta_i$ in the score function $\bm \epsilon_{\bm \theta}(\bm z_s,t_s)$, the task-specific sensitivity $g_i$ is approximated by accumulating squared gradients over $N$ iterations:
\begin{equation}
\label{eq:accum}
    g_i = \sum_{j=1}^N\left(\frac{\partial\ell(\bm \theta)}{\partial\theta^{(j)}_i}\right)^2,
\end{equation}
where $i$ and $j$ are the parameter and iteration indices, respectively. This metric identifies parameters most responsive to the hiding objective.

\noindent\textbf{Sensitive Layer Selection.} We extend parameter-level to layer-level sensitivity through spatial aggregation. Specifically, we begin by binarizing the sensitivity score of each parameter $g_i$:
\begin{equation}
\label{eq:threshold}
b_i= \begin{cases}1 & g_i \ge \tau  \\ 0 & g_i < \tau \end{cases},
\end{equation}
where $\tau$ is a predefined threshold that controls the sensitive parameter sparsity. Next, we rank all network layers by their counts of sensitive parameters (\ie, those with $b_i = 1$), and select top-$\eta$ layers for subsequent structured and computationally efficient fine-tuning. 
Critically, we avoid summing raw sensitivity scores within layers, as this could bias selection towards layers with disproportionately many parameters---even if those parameters are less impactful (known as the law of triviality). By prioritizing the number of sensitive parameters over their cumulative sensitivity, we ensure a balanced layer-wise assessment.

\noindent\textbf{LoRA-based Fine-tuning.} For each selected sensitive layer, we implement a variant of LoRA for PEFT. Specifically, for linear layers, we apply the standard LoRA~\cite{hu2022lora} by injecting trainable low-rank matrices $\Delta 
\mathbf W = \mathbf A\mathbf B$, where $\mathbf A \in \mathbb{R}^{m\times r}$ and $\mathbf B \in \mathbb{R}^{r\times n}$ with rank $r \ll \min\{m,n\}$. For convolutional layers, we first reshape the 4D convolution filters into 2D matrices by flattening the input channel and filter dimensions, and then apply LoRA accordingly. To improve fine-tuning performance and accelerate convergence, we incorporate the rank stabilization trick proposed in~\cite{kalajdzievski2023rslora}, adjusting the scaling factor to be proportional to $O(1 / \sqrt{r})$. Meanwhile, we apply the learning rate decoupling trick introduced by~\cite{hayou2024lora+}, setting different learning rates for the matrices $\mathbf{A}$ and $\mathbf{B}$, respectively.

\begin{figure*}[t]
  \centering
  \includegraphics[width=\textwidth]{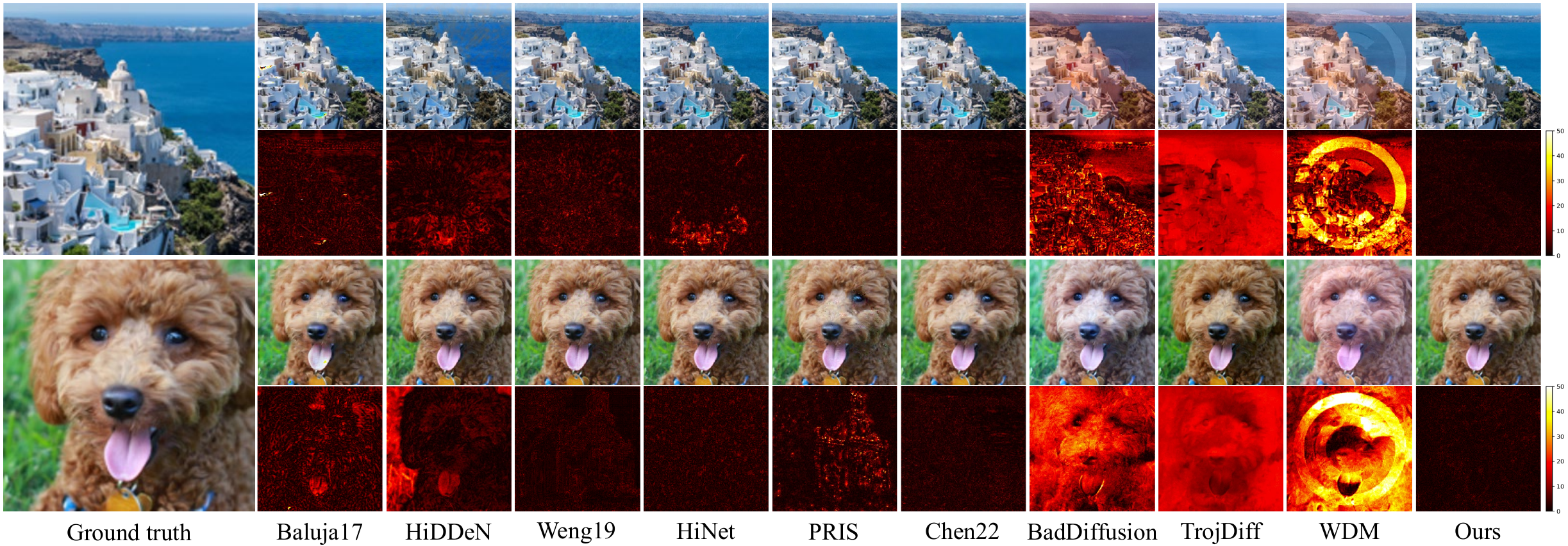}
  \caption{Visual comparison of extracted secret images along with the absolute error maps.}
  \label{fig:fd}
\end{figure*}

\begin{figure*}[t]
  \centering
    \subfloat[Original]{\includegraphics[width=0.198\textwidth]{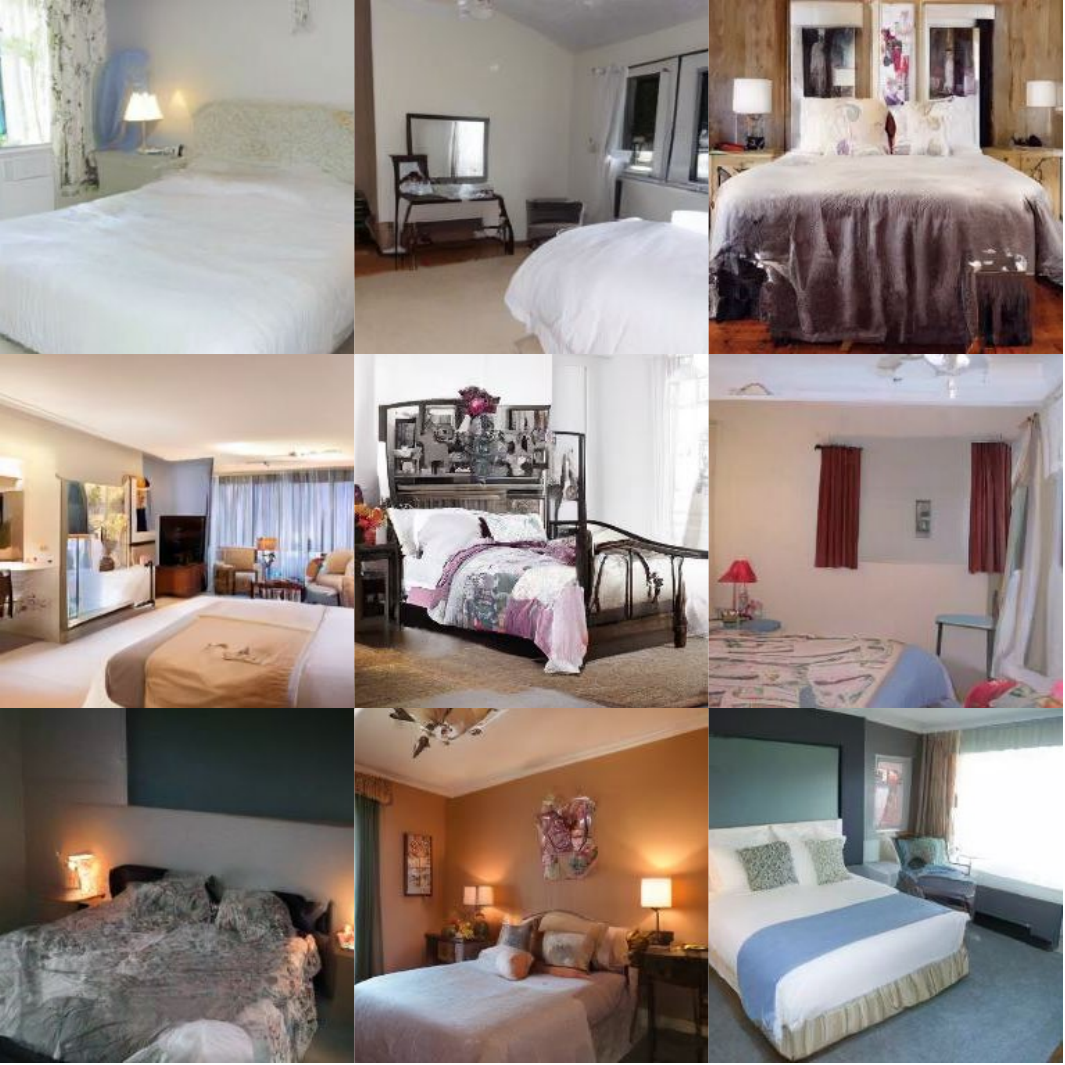}}
    \hfill
    \subfloat[BadDiffusion]{\includegraphics[width=0.198\textwidth]{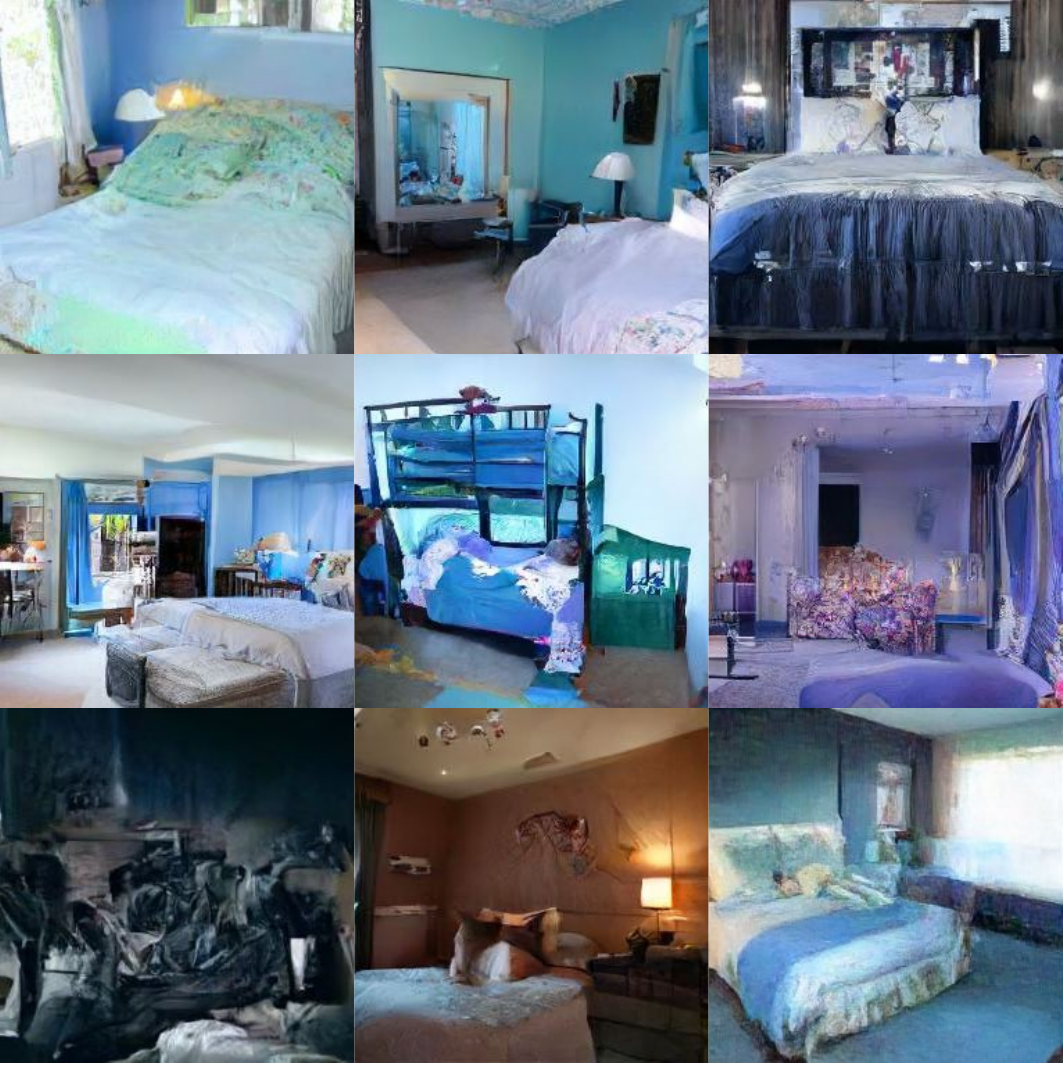}} 
    \hfill
    \subfloat[TrojDiff]{\includegraphics[width=0.198\textwidth]{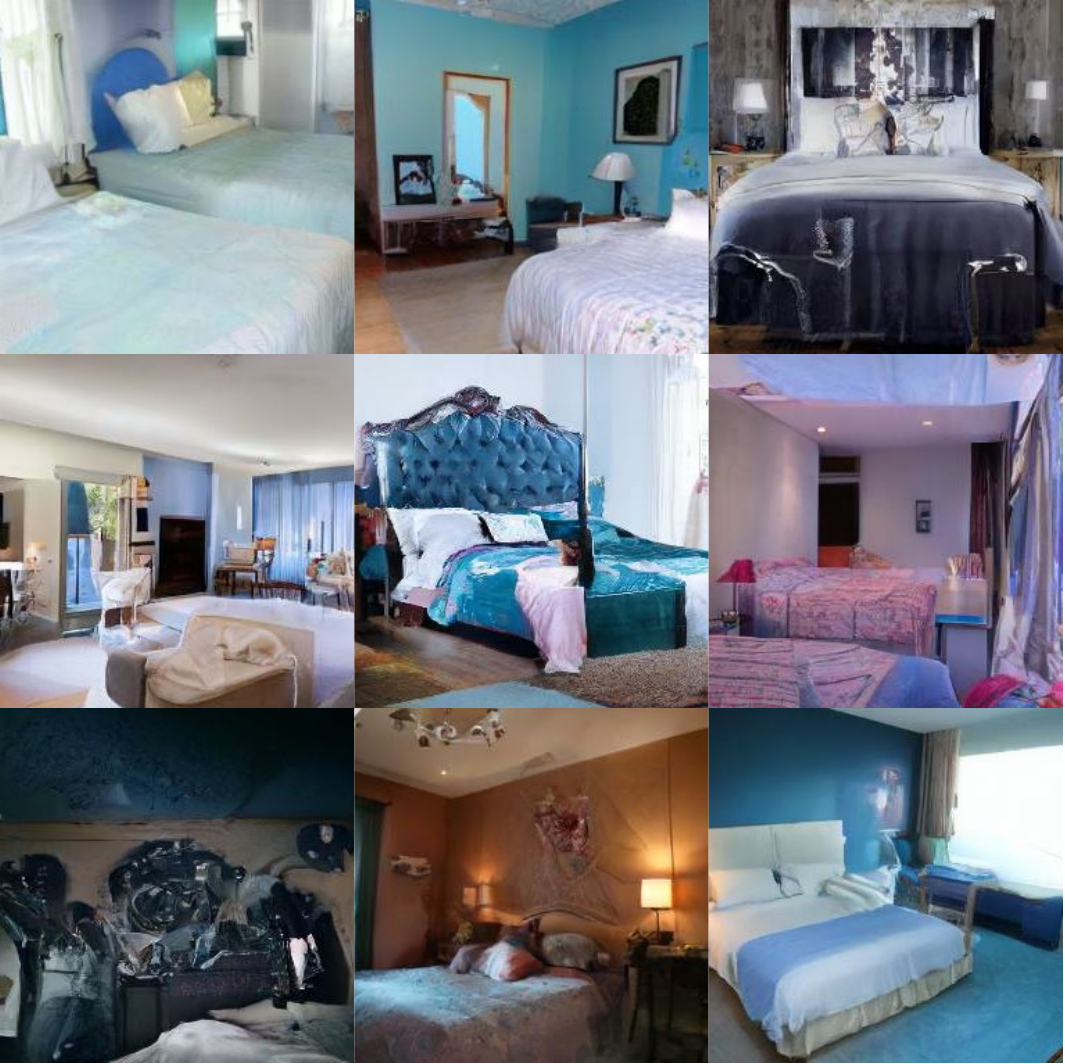}} 
    \hfill
    \subfloat[WDM]{\includegraphics[width=0.198\textwidth]{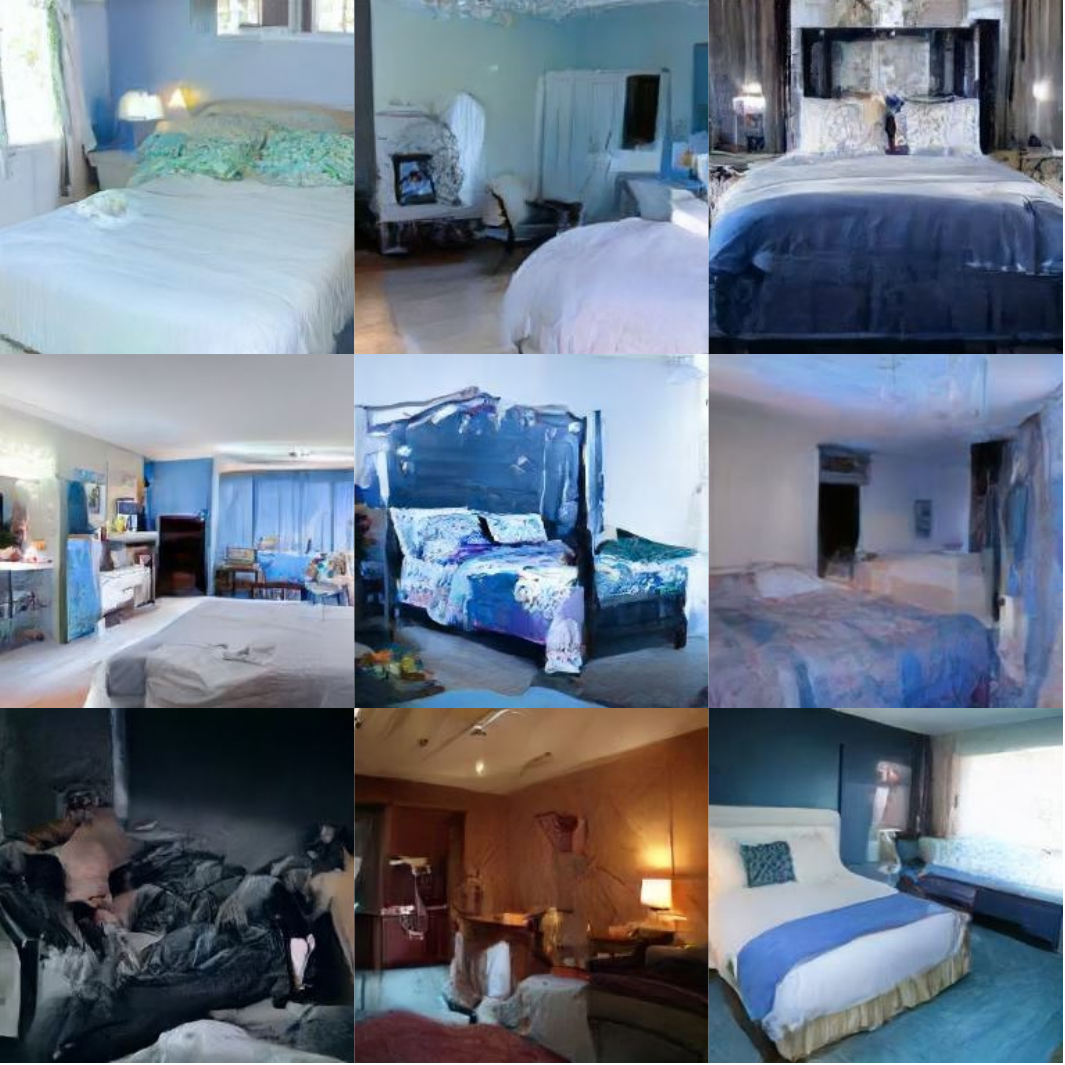}} 
    \hfill
    \subfloat[Ours]{\includegraphics[width=0.198\textwidth]{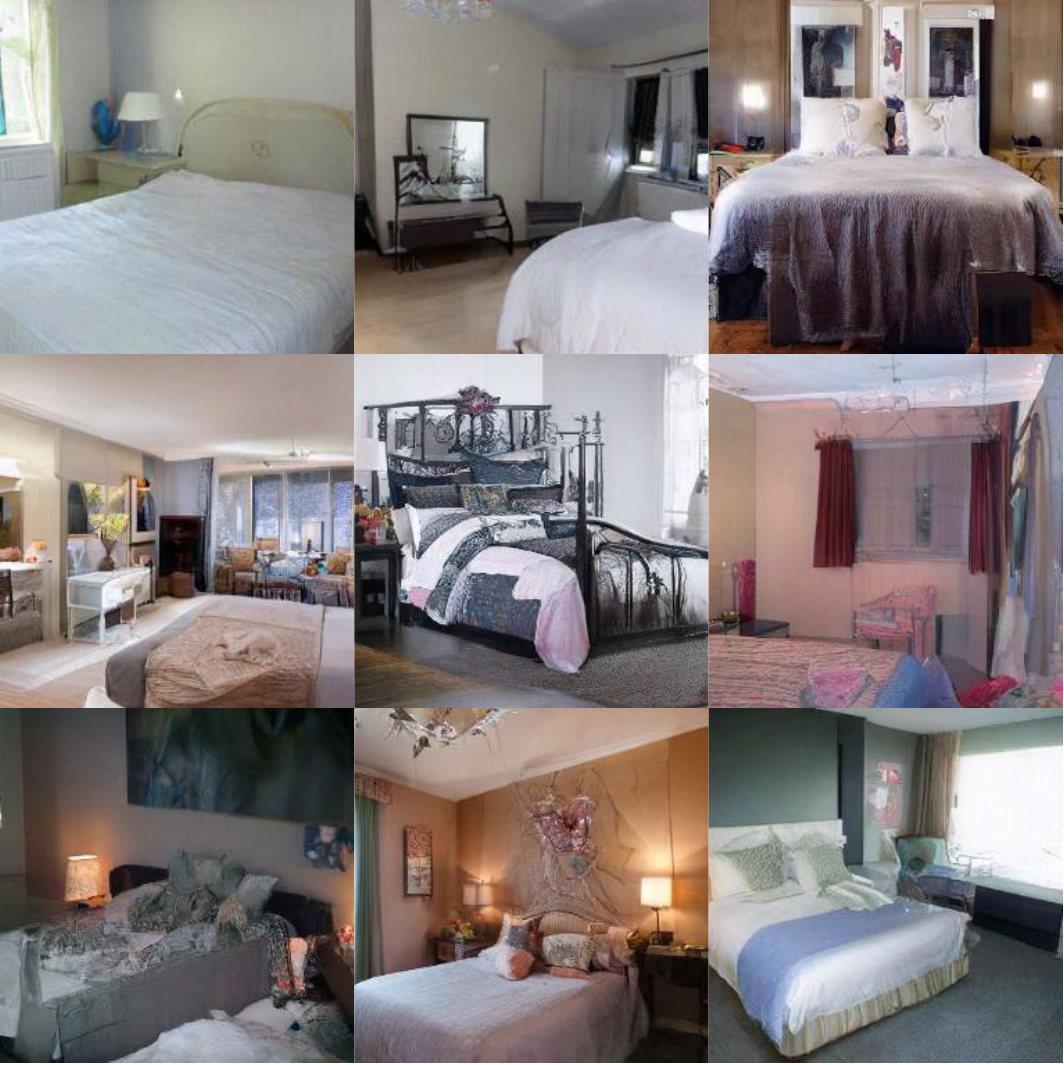}} 
  \caption{Visual comparison of images generated by the original and stego diffusion models with the same set of initial noise.}
  \label{fig:asd}
\end{figure*}

\section{Experiments}
\label{sec:exp}
In this section, we first describe the experimental setups. Subsequently, we present comprehensive comparisons against nine state-of-the-art neural steganography methods across three key dimensions: extraction accuracy, model fidelity, and hiding efficiency. Finally, we extend our analysis to multi-image hiding scenarios, demonstrating the scalability of our approach.

\subsection{Experimental Setups}
\noindent\textbf{Implementation Details.} 
We employ the pre-trained DDPMs~\cite{ho2020denoising} as our default pixel-space diffusion models at two image resolutions: $32\times 32$ pixels using CIFAR10~\cite{krizhevsky2009learning} and $256\times 256$ pixels using LSUN bedroom~\cite{yu2015lsun}. Unlike previous approaches~\cite{chou2023backdoor, peng2023protecting, chen2023trojdiff}, which typically embed overly simplistic image patterns like QR codes or icons, we assess our method with complex natural images aggregated from widely used datasets, including COCO~\cite{lin2014microsoft}, DIV2K~\cite{agustsson2017div2k}, LSUN Church~\cite{yu2015lsun}, and Places~\cite{zhou2018places}.

The architectures and hyperparameters of the DDPMs follow the specifications outlined in~\cite{ho2020denoising}. By default, secret embedding and extraction are performed at timestep $t_s = 500$. For parameter sensitivity analysis, we accumulate gradients over $N=50$ iterations. The threshold in Eq.~\eqref{eq:threshold} is selected to achieve sensitive parameter sparsity levels of $0.01$ for $32\times 32$ images and $0.1$ for $256\times 256$ images. Correspondingly, the number of sensitive layers $\eta$ is set to $15$ for $32\times32$ images and $45$ for $256\times256$ images, respectively. The maximum iteration number for LoRA is set to $2,000$, with rank $r = 64$ for $32\times 32$ images and $r=128$ for $256\times256$ images.

\noindent\textbf{Evaluation Criteria.} Our evaluation encompasses three aspects: 1) extraction accuracy, quantified via PSNR, the structural similarity (SSIM) index~\cite{wang2004image}, the learned perceptual image patch similarity (LPIPS) measure~\cite{zhang2018perceptual}, and the deep image structure and texture similarity (DISTS) metric~\cite{ding2020image} between the original and extracted secret images; 2) model fidelity, assessed at both the sample-level using PSNR, SSIM, LPIPS and DISTS, and the population-level using FID~\cite{heusel2017gans} to measure deviations between generated images from the original and stego diffusion models; and 3) hiding efficiency, measured by GPU hours required for embedding a secret image on an AMD EPYC 7F52 16-Core CPU and NVIDIA GeForce RTX3090 GPUs.

\subsection{Main Results}
We compare our method with five autoencoder-based algorithms: 1) Baluja17~\cite{baluja2017hiding}, 2) HiDDeN~\cite{zhu2018hiding}, 3) Weng19~\cite{weng2019high}, 4) HiNet~\cite{jing2021hinet} and 5) PRIS~\cite{yang2024pris}, one GAN-based technique: 6) Chen22~\cite{chen2022hiding}, and three diffusion-based methods: 7) BadDiffusion~\cite{chou2023backdoor}, 8) TrojDiff~\cite{chen2023trojdiff} and 9) WDM~\cite{peng2023protecting}. 

\noindent\textbf{Extraction Accuracy.} 
As shown in Table~\ref{tab:tf}, autoencoder-based techniques~\cite{baluja2017hiding, zhu2018hiding, weng2019high, jing2021hinet, yang2024pris} achieve higher extraction accuracy at higher image resolutions. In contrast, existing GAN-based~\cite{chen2022hiding} and diffusion-based~\cite{chou2023backdoor,peng2023protecting, chen2023trojdiff} methods perform better at lower image resolutions. Moreover, our approach outperforms all existing neural steganography methods across both image resolutions. Fig.~\ref{fig:fd} presents a qualitative comparison of extraction accuracy using absolute error maps between the original and extracted secret images at the resolution of $256\times256$. The visual results indicate that our method produces the least reconstruction errors across all spatial locations. 

\noindent\textbf{Model Fidelity.} 
Table~\ref{tab:asd} shows that our method maintains the closest model fidelity to the original DDPMs, as evidenced by minimal FID variations ($4.77$ versus $4.79$ for $32\times 32$ images and $8.39$ versus $7.46$ for $256\times 256$ images). Sample-level fidelity metrics (\ie, PSNR, SSIM, LPIPS, and DISTS) also validate the ability of our method to produce visually indistinguishable outputs from the original model. 
Fig.~\ref{fig:asd} compares the $256\times256$ images generated by the original and different stego diffusion models, using the same set of initial noise. It is clear that our method produces images that closely resemble those from the original model, whereas other stego diffusion models introduce noticeable discrepancies in structural details and color appearances of objects and backgrounds.

\noindent\textbf{Hiding Efficiency.}
As also reported in Table~\ref{tab:asd}, our method significantly reduces the hiding time to only $0.04$ GPU hours for $32\times 32$ images and $0.18$ GPU hours for $256\times 256$ images, making it over $50$ times faster than the second best method, WDM~\cite{peng2023protecting}. This substantial efficiency improvement arises because we embed secret images only at a privately chosen timestep $t_s$, rather than fine-tuning the entire reverse diffusion process over numerous iterations.

\noindent\textbf{Hiding Multiple Images.}
To explore scalability, we further assess our method in scenarios involving multiple secret images intended for different recipients. Table~\ref{tab:mul} demonstrates that although embedding additional images slightly sacrifices model fidelity, our method maintains high extraction accuracy, supporting secure communication among multiple recipients.

In summary, our comprehensive experiments confirm that the proposed method achieves superior extraction accuracy, high model fidelity, and exceptional hiding efficiency, establishing a new benchmark in diffusion-based neural steganography.

\begin{table}[t]
  \caption{Extraction accuracy and model fidelity of our method when hiding multiple secret images for different recipients at two image resolutions.}

  \label{tab:mul}
  \centering
  \resizebox{0.95\linewidth}{!}{\begin{tabular}{lccccc}
    \toprule
  Measure &$\#$&{PSNR$\uparrow$}   &{SSIM$\uparrow$}   &{LPIPS$\downarrow$}  &{DISTS$\downarrow$}         
    \\
    \midrule
    \multicolumn{2}{l}{\cellcolor{lightgray!20}{\textit{$32\times32$}}} &\cellcolor{lightgray!20} &\cellcolor{lightgray!20} &\cellcolor{lightgray!20} &\cellcolor{lightgray!20}\\
    \multirow{3}{*}{{Extraction Accuracy}} 
    &1                      & $52.90$  & $0.99$  &$0.001$ & $0.001$\\
    &2                      & $50.44$  & $0.99$  &$0.001$ & $0.001$\\
    &4                      & $49.38$  & $0.99$  &$0.001$ & $0.001$\\
    \midrule
    \multirow{3}{*}{{Model Fidelity}}
    &1                      & $31.06$     & $0.94$  &$0.077$ & $0.037$\\
    &2                      & $30.86$     & $0.95$  &$0.067$ & $0.035$\\
    &4                      & $30.93$     & $0.95$  &$0.064$ & $0.033$\\
    \midrule
    \multicolumn{2}{l}{\cellcolor{lightgray!20}{\textit{$256\times256$}}} &\cellcolor{lightgray!20} &\cellcolor{lightgray!20} &\cellcolor{lightgray!20} &\cellcolor{lightgray!20}\\
    \multirow{3}{*}{{Extraction Accuracy}} 
    &1                    & $39.33$  & $0.97$  &$0.043$ & $0.018$\\
    &2                    & $38.61$  & $0.97$  &$0.034$ & $0.023$\\
    &4                    & $38.31$  & $0.96$  &$0.058$ & $0.029$\\
    \midrule
    \multirow{3}{*}{{Model Fidelity}}
    &1                    & $23.31$  & $0.83$  &$0.235$ & $0.112$\\
    &2                    & $20.53$  & $0.78$  &$0.313$ & $0.137$\\
    &4                    & $17.78$  & $0.74$  &$0.394$ & $0.165$\\
    \bottomrule
  \end{tabular}}
\end{table}

\begin{table}[t]
    \caption{Ablation on the number of iterations for parameter sensitivity accumulation.}
    \label{tab:abl-n}
    \centering
    \resizebox{0.9\linewidth}{!}{\begin{tabular}{lcccc}
    \toprule
    {$\#$ of Iterations} &\multicolumn{1}{c}{PSNR$\uparrow$}   &\multicolumn{1}{c}{SSIM$\uparrow$}   &\multicolumn{1}{c}{LPIPS$\downarrow$}  &\multicolumn{1}{c}{DISTS$\downarrow$}         
    \\
    \midrule
    \cellcolor{lightgray!20}{\textit{Extraction Accuracy}} &\cellcolor{lightgray!20}&\cellcolor{lightgray!20}&\cellcolor{lightgray!20}&\cellcolor{lightgray!20}\\
    $1$     & $40.93$  & $0.99$ & $0.001$ & $0.006$ \\
    $50$    & $52.90$  & $0.99$ & $0.001$ & $0.001$ \\
    $100$   & $53.08$  & $0.99$ & $0.001$ & $0.001$ \\
    
    \midrule
    \cellcolor{lightgray!20}{\textit{Model Fidelity}} &\cellcolor{lightgray!20}&\cellcolor{lightgray!20}&\cellcolor{lightgray!20}&\cellcolor{lightgray!20}\\ 
    $1$     & $28.58$ & $0.92$ & $0.101$ & $0.046$ \\
    $50$    & $31.06$ & $0.94$ & $0.077$ & $0.037$ \\
    $100$   & $31.20$ & $0.94$ & $0.074$ & $0.036$ \\
    
    \bottomrule
    \end{tabular}}
\end{table}

\begin{table}[t]
    \caption{Ablation on the number of selected sensitive layers.}
    \label{tab:abl-layers}
    \centering
    \resizebox{0.9\linewidth}{!}{\begin{tabular}{lcccc}
    \toprule
    {$\#$ of Sensitive Layers} &\multicolumn{1}{c}{PSNR$\uparrow$}   &\multicolumn{1}{c}{SSIM$\uparrow$}   &\multicolumn{1}{c}{LPIPS$\downarrow$}  &\multicolumn{1}{c}{DISTS$\downarrow$}         
    \\
    \midrule
    \cellcolor{lightgray!20}{\textit{Extraction Accuracy}} &\cellcolor{lightgray!20}&\cellcolor{lightgray!20}&\cellcolor{lightgray!20}&\cellcolor{lightgray!20}\\
        $5$ & $47.48$ & $0.99$ & $0.001$ & $0.002$ \\
        $15$& $52.90$ & $0.99$ & $0.001$ & $0.001$  \\
        $45$& $54.04$ & $0.99$ & $0.001$ & $0.001$  \\
    \midrule
    \cellcolor{lightgray!20}{\textit{Model Fidelity}} &\cellcolor{lightgray!20}&\cellcolor{lightgray!20}&\cellcolor{lightgray!20}&\cellcolor{lightgray!20}\\ 
        $5$ & $31.55$ & $0.95$ & $0.068$ & $0.035$ \\
        $15$& $31.06$ & $0.94$ & $0.077$ & $0.037$ \\
        $45$& $27.87$ & $0.91$ & $0.122$ & $0.048$ \\
    \bottomrule
    \end{tabular}}
\end{table}

\subsection{Ablation Studies}
\label{sec:abl}
To systematically analyze the contributions of key components in the proposed method, we conduct comprehensive ablation studies on $32\times 32$ images.

\noindent\textbf{Parameter Sensitivity Accumulation.}
We first evaluate how the number of iterations $N$, used for sensitivity accumulation in Eq.~\eqref{eq:accum} impacts performance. As detailed in Table~\ref{tab:abl-n}, we observe significant performance gains when increasing $N$ from $1$ to $50$ in terms of both extraction accuracy and model fidelity. However, further increasing $N$ to $100$ yields negligible benefits relative to the computational overhead incurred. Consequently, we set  $N=50$ as a reasonable trade-off between method performance and computational efficiency.

\noindent\textbf{Number of Selected Layers.} We next analyze how the selection of different numbers of sensitive layers affects performance. As demonstrated in Table~\ref{tab:abl-layers}, selecting too few layers (\eg, $5$) results in inadequate extraction accuracy (PSNR of $47.48$), despite good model fidelity. Conversely, selecting too many layers (\eg, $45$) reduces model fidelity due to excessive parameter adjustments. A balanced selection of $15$ layers significantly enhances extraction accuracy (PSNR of $52.90$) without compromising model fidelity, and thus is adopted as the default choice.

\begin{figure}[t]
  \centering
  \includegraphics[width=\linewidth]{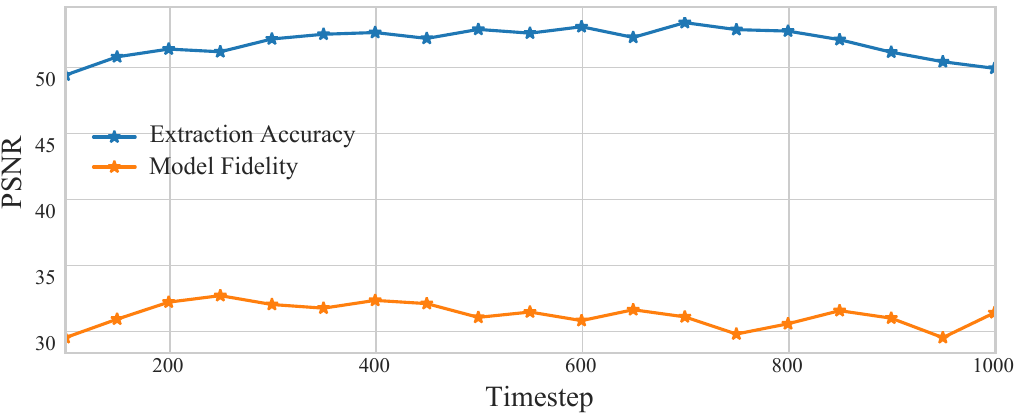}
  \caption{Ablation on the selection of the secret timestep.}
  \label{fig:abl-t}
\end{figure}

\noindent\textbf{Secret Timestep Selection.}
The proposed method introduces flexibility in the choice of the secret embedding timestep $t_s$. Fig.~\ref{fig:abl-t} illustrates how varying the choice of $t_s$ affects performance. Encouragingly, extraction accuracy and model fidelity remain consistently robust across a broad range of timesteps, underscoring the practical advantage and reliability of our approach in various operational (\eg, multi-image and multi-recipient) scenarios.

\noindent\textbf{Number of PEFT Iterations.}
We further explore how varying the number of iterations in our PEFT method influences performance. As shown in Table~\ref{tab:iters}, increasing the number of PEFT iterations generally improves extraction accuracy but reduces model fidelity and substantially increases computational cost, as expected. For example, with fewer iterations (\eg, $1,000$), our method yields high model fidelity but suffers from inadequate extraction accuracy.  Thus, we choose $2,000$ iterations as our default configuration, as it achieves a balanced trade-off between extraction accuracy, model fidelity, and hiding efficiency.

\noindent\textbf{PEFT versus Full Fine-tuning.}
We compare our PEFT strategy against full fine-tuning to evaluate its effectiveness and efficiency. The results in Table~\ref{tab:fft}  clearly demonstrate that our PEFT method significantly outperforms full fine-tuning in maintaining model fidelity. Notably, this superior fidelity is achieved with minimal sacrifice in extraction accuracy and a significant reduction in computational time---requiring only half the time of full fine-tuning.

\noindent\textbf{Robustness to Image Noise.} 
To assess robustness against potential input noise, we intentionally perturb secret natural images with random Gaussian noise (with mean zero and variance $100$) before embedding. Table~\ref{tab:fft} confirms the resilience of our method, showing minimal performance degradation under substantial noise conditions. 
This verifies the potential applicability of our method in encrypted transmission environments.

\noindent\textbf{Compatibility Across Diffusion Model Variants.}
Lastly, we validate the adaptability of our method by testing it across several popular diffusion model variants, including DDPM~\cite{ho2020denoising}, DDPM++~\cite{song2021score}, DDIM~\cite{song2021denoising}, and EDM~\cite{karras2022elucidating}. Table~\ref{tab:type} demonstrates that high extraction accuracy and model fidelity are maintained across these diverse diffusion model architectures, underscoring the broad generalizability and versatility of our method. Note that during DDIM inference, certain denoising steps may be skipped, and this skipping might inadvertently include the selected secret timestep, which explains the best model fidelity.

Overall, these ablation studies collectively validate the efficiency, robustness, and adaptability of our method, firmly establishing its practical effectiveness in neural steganography applications.

\begin{table}[t]
  \caption{Ablation on the number of PEFT iterations.}
  \label{tab:iters}
  \centering
  \resizebox{\linewidth}{!}{\begin{tabular}{lccccccc}
    \toprule
    {$\#$ of Iterations} &\multicolumn{1}{c}{PSNR$\uparrow$}   &\multicolumn{1}{c}{SSIM$\uparrow$}   &\multicolumn{1}{c}{LPIPS$\downarrow$}  &\multicolumn{1}{c}{DISTS$\downarrow$} & Time$\downarrow$        
    \\
    \midrule
    \cellcolor{lightgray!20}{\textit{Extraction Accuracy}} &\cellcolor{lightgray!20}&\cellcolor{lightgray!20}&\cellcolor{lightgray!20}&\cellcolor{lightgray!20}&\cellcolor{lightgray!20}\\
    $1,000$    & $49.31$ & $0.99$ & $0.001$ & $0.001$ & $0.02$\\
    $2,000$    & $52.90$ & $0.99$ & $0.001$ & $0.001$ & $0.04$\\
    $5,000$    & $55.16$ & $0.99$ & $0.001$ & $0.001$ & $0.11$\\
    \midrule
    \cellcolor{lightgray!20}{\textit{Model Fidelity}} &\cellcolor{lightgray!20}&\cellcolor{lightgray!20}&\cellcolor{lightgray!20}&\cellcolor{lightgray!20}&\cellcolor{lightgray!20}\\ 
    $1,000$    & $32.54$ & $0.96$ & $0.067$ & $0.032$ &--\\
    $2,000$    & $31.06$ & $0.94$ & $0.077$ & $0.037$ &--\\
    $5,000$    & $25.88$ & $0.91$ & $0.137$ & $0.057$ &--\\
    \bottomrule
    \end{tabular}}
\end{table}

\begin{table}[t]
  \caption{Ablations on PEFT against full fine-tuning, and on hiding clean and noisy natural images.}

  \label{tab:fft}
  \centering
  \resizebox{\linewidth}{!}{\begin{tabular}{lccccccc}
    \toprule
    {Method} &\multicolumn{1}{c}{PSNR$\uparrow$}   &\multicolumn{1}{c}{SSIM$\uparrow$}   &\multicolumn{1}{c}{LPIPS$\downarrow$}  &\multicolumn{1}{c}{DISTS$\downarrow$} & Time$\downarrow$        
    \\
    \midrule
    \cellcolor{lightgray!20}{\textit{Extraction Accuracy}} &\cellcolor{lightgray!20}&\cellcolor{lightgray!20}&\cellcolor{lightgray!20}&\cellcolor{lightgray!20}&\cellcolor{lightgray!20}\\
    Full Fine-tuning & $53.55$   & $0.99$  &$0.001$ & $0.001$ &$0.08$\\
    Hiding Noisy Image  & $51.86$   & $0.99$   & $0.001$  & $0.001$ &$0.04$\\
    Ours        & $52.90$  & $0.99$   &$0.001$ & $0.001$ &$0.04$\\
    \midrule
    \cellcolor{lightgray!20}{\textit{Model Fidelity}} &\cellcolor{lightgray!20}&\cellcolor{lightgray!20}&\cellcolor{lightgray!20}&\cellcolor{lightgray!20}&\cellcolor{lightgray!20}\\ 
    Full Fine-tuning & $23.98$    &$0.84$  &$0.212$ & $0.075$ &--\\
    Hiding Noisy Image  & $31.30$  & $0.94$  & $0.072$  & $0.035$ &--\\
    Ours       & $31.06$    & $0.94$ &$0.077$ & $0.037$ &--\\
    \bottomrule
  \end{tabular}}
\end{table}

\begin{table}[t]
  \caption{Ablation on different diffusion models. Unlike DDPM, DDPM++, and DDIM, the EDM variant is pre-trained on the larger-sized ImageNet dataset ($64\times64$), explaining its slightly lower extraction accuracy and reduced model fidelity.}

  \label{tab:type}
  \centering
  \resizebox{0.9\linewidth}{!}{\begin{tabular}{lccccccc}
    \toprule
    {Diffusion Model} &{PSNR$\uparrow$} &{SSIM$\uparrow$} &{LPIPS$\downarrow$} &{DISTS$\downarrow$}  \\
    \midrule
    \cellcolor{lightgray!20}{\textit{Extraction Accuracy}} &\cellcolor{lightgray!20}&\cellcolor{lightgray!20}&\cellcolor{lightgray!20}&\cellcolor{lightgray!20}\\
    DDPM~\cite{ho2020denoising}   & $52.90$   & $0.99$  &$0.001$  & $0.001$\\
    DDPM++~\cite{song2021score} & $53.05$   & $0.99$  & $0.001$ & $0.001$\\
    DDIM~\cite{song2021denoising}   & $52.96$   & $0.99$  & $0.001$ & $0.001$\\
    EDM~\cite{karras2022elucidating}    & $51.30$   & $0.99$  & $0.002$ & $0.001$\\
    \midrule
    \cellcolor{lightgray!20}{\textit{Model Fidelity}} &\cellcolor{lightgray!20}&\cellcolor{lightgray!20}&\cellcolor{lightgray!20}&\cellcolor{lightgray!20}\\ 
    DDPM~\cite{ho2020denoising}     & $31.06$    & $0.94$  &$0.077$  & $0.037$\\
    DDPM++~\cite{song2021score}  & $31.37$    & $0.95$  & $0.073$ & $0.035$\\
    DDIM~\cite{song2021denoising}    & $35.08$    & $0.97$  & $0.034$ & $0.022$\\
    EDM~\cite{karras2022elucidating}     & $28.80$    & $0.94$  & $0.090$ & $0.056$\\
    \bottomrule
  \end{tabular}}
\end{table}

\section{Conclusion and Discussion}
We have proposed a computational method for hiding images in diffusion models by editing learned score functions at specific timesteps, through a hybrid PEFT strategy. Our method achieves outstanding extraction accuracy, preserves model fidelity, and significantly enhances hiding efficiency compared to existing methods. Through extensive experiments and rigorous ablation studies, we highlighted critical factors contributing to the superior performance of our method. The localized score function editing turns out to be a notable contribution, enabling precise adjustments with minimal computational cost.

Looking ahead, several promising future research directions emerge from this study. One particularly compelling avenue is the development of adaptive hiding strategies that dynamically adjust embedding parameters based on data characteristics or security requirements. Moreover, rigorous theoretical analysis of diffusion model behaviors under various embedding scenarios, including theoretical upper bounds on detectability, will be critical to fully exploit and understand the potential of diffusion-based neural steganography. Lastly, improving embedding robustness against model perturbations---such as noising, pruning, and compression---could transform our method into a highly effective digital watermarking solution, facilitating authentication and copyright protection of diffusion models.

\section*{Acknowledgments}
This work was supported in part by the Hong Kong
RGC General Research Fund (11220224) and the Hong Kong ITC
Innovation and Technology Fund (9440390).

{
    \small
    \bibliographystyle{ieeenat_fullname}
    \bibliography{main}
}

\clearpage
\setcounter{page}{1}

\onecolumn
\section*{Appendix}
\renewcommand{\thesubsection}{A1}
\subsection{More Visual Examples}
Fig.~\ref{fig:acc-2s} show visual examples of extracted secret images along with absolute error maps from our stego diffusion models, when hiding two and four images, respectively. 
Fig.~\ref{fig:modl-fd} compares images generated by the original and stego diffusion models when hiding two and four secret images, respectively.

\renewcommand{\thefigure}{A1}
\begin{figure*}[h!]
    \centering
    \subfloat[]{\includegraphics[width=0.49\linewidth]{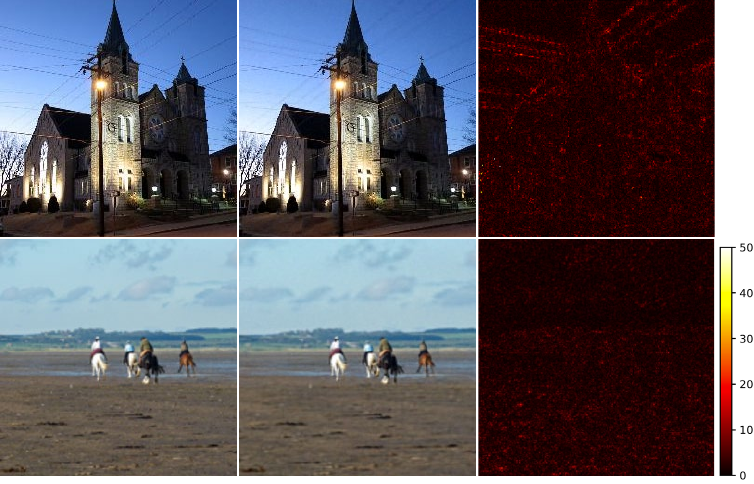}}
    \hfill
    \subfloat[]{\includegraphics[width=0.49\linewidth]{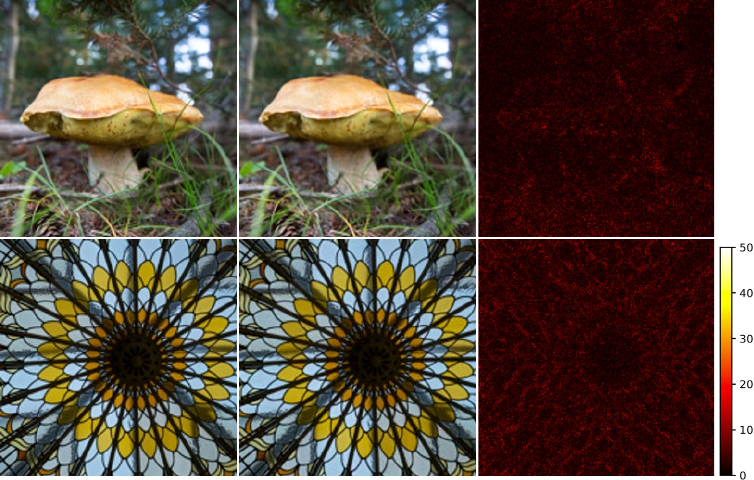}}
    \vfill
    \subfloat[]{\includegraphics[width=0.49\linewidth]{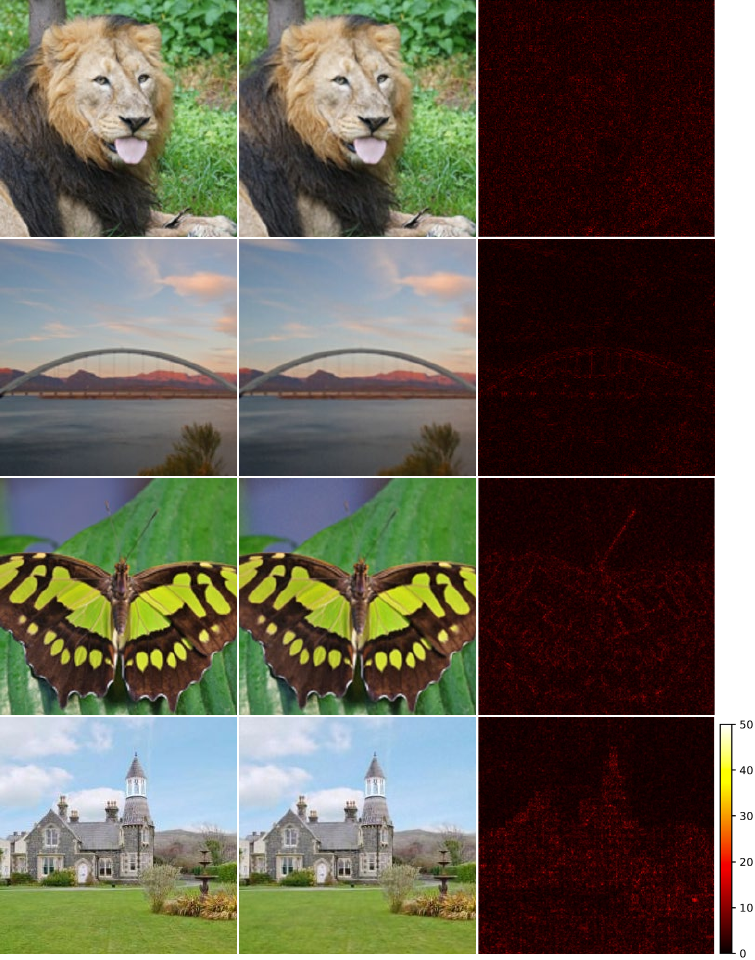}}
    \hfill
    \subfloat[]{\includegraphics[width=0.49\linewidth]{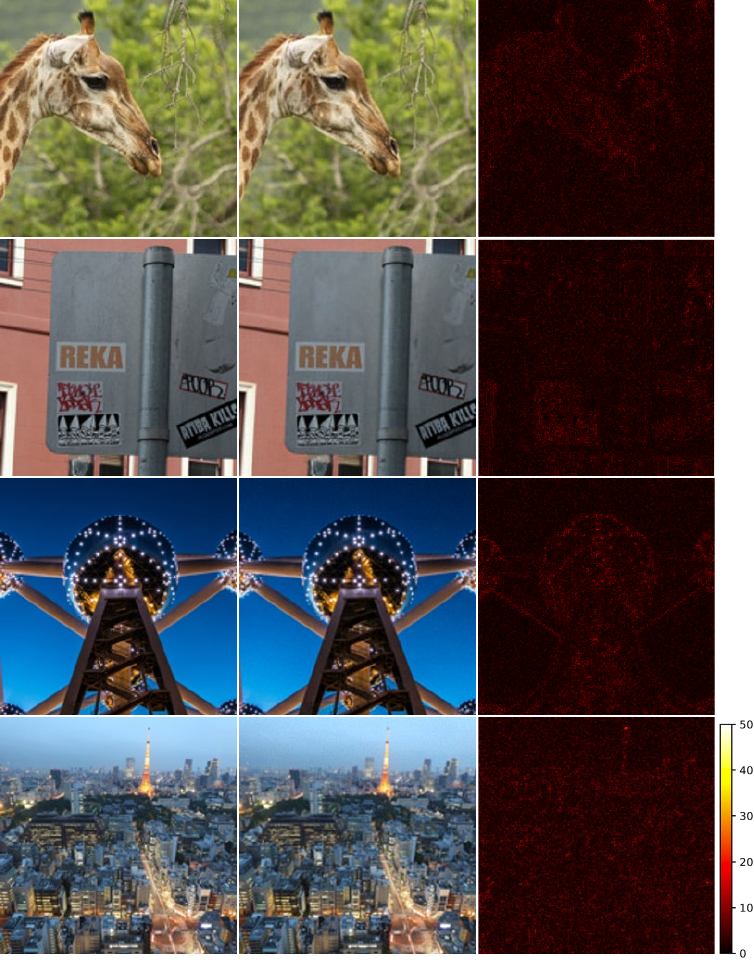}}

    \caption{Visual examples of extracted secret images and corresponding absolute error maps for hiding two images (subfigures (a) and (b)) and four images (subfigures (c) and (d)). In each subfigure, columns from left to right show the ground-truth secret images, extracted secret images, and absolute error maps, respectively.}
    \label{fig:acc-2s}
\end{figure*}

\renewcommand{\thefigure}{A2}
\begin{figure*}[t!]
  \centering
    \subfloat[Original]{\includegraphics[width=0.8\textwidth]{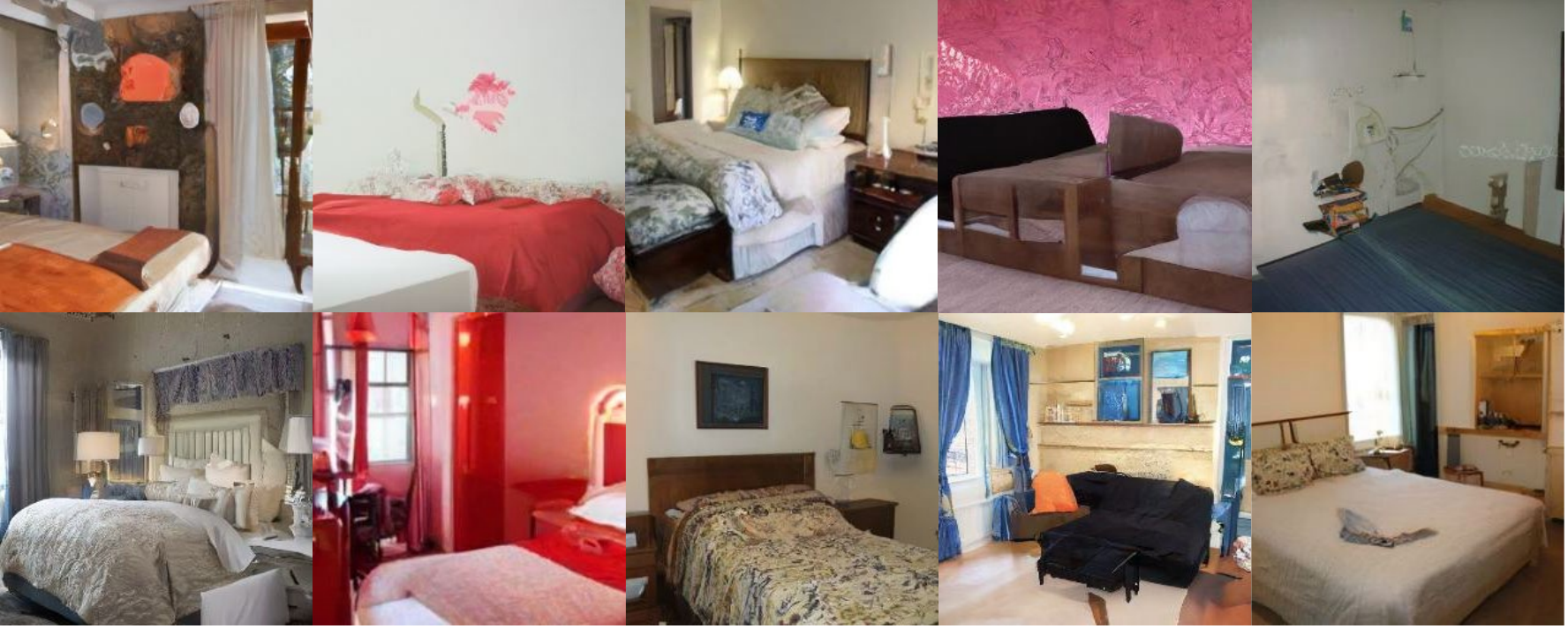}}
    \vfill
    \subfloat[Stego model for hiding two images]{\includegraphics[width=0.8\textwidth]{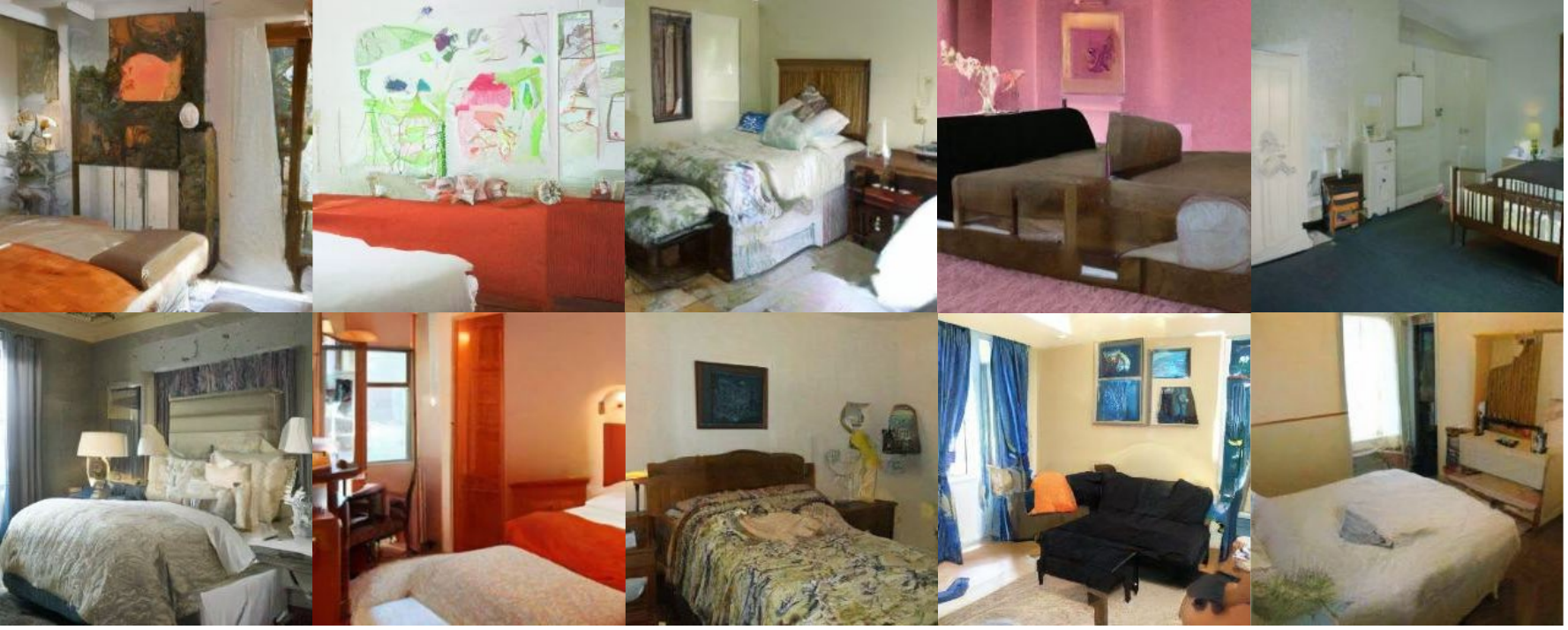}} 
    \vfill
    \subfloat[Stego model for hiding four images]{\includegraphics[width=0.8\textwidth]{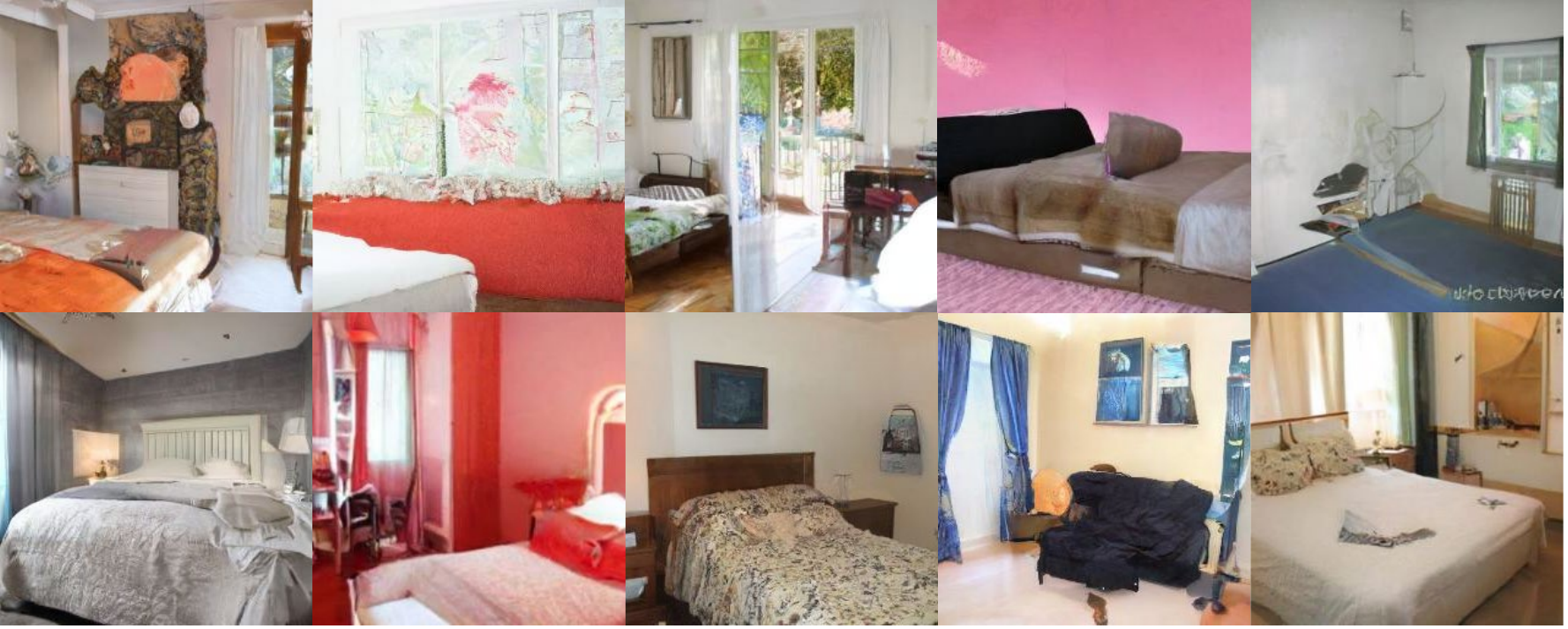}} 
  \caption{Visual comparison of images generated by the original and stego diffusion models embedding two and four secret images.}
  \label{fig:modl-fd}
\end{figure*}

\end{document}